\documentclass[conference]{IEEEtran}
\usepackage{times}

\usepackage[numbers]{natbib}
\usepackage{multicol}
\usepackage[bookmarks=true]{hyperref}
\usepackage{authblk}
\makeatletter
\renewcommand\AB@affilsepx{\quad\protect\Affilfont}

\makeatother
\IEEEoverridecommandlockouts

\usepackage[table]{xcolor}
\usepackage{amsmath}
\usepackage{amsfonts}
\usepackage{amssymb}
\usepackage{mathtools}
\usepackage{dsfont}
\usepackage{graphicx}
\usepackage{booktabs}
\usepackage{caption}
\usepackage{subcaption}
\usepackage{hyperref}
\usepackage{cleveref}
\usepackage{xfrac}
\usepackage{cancel}
\usepackage[thinc]{esdiff}
\usepackage{siunitx}
\usepackage[normalem]{ulem}
\usepackage{svg}
\usepackage{etoolbox}
\usepackage{comment}
\usepackage[linesnumbered, ruled, vlined]{algorithm2e}
\usepackage{ulem}
\usepackage{multirow}

\title{Model Predictive Control for Aggressive Driving Over Uneven Terrain}

\author{Tyler Han$^*$\thanks{*Corresponding Author: than123@cs.washington.edu}}
\author{Alex Liu}
\author{Anqi Li}
\author{Alexander Spitzer}
\author{Guanya Shi}
\author{Byron Boots\vspace{-6pt}}
\affil{University of Washington\vspace{-12pt}}
\affil{%
\textcolor{red}{%
\url{https://sites.google.com/cs.washington.edu/off-road-mpc}}\vspace{-6pt}%
}

\begin{document}

\urlstyle{tt}

\newcommand{\add}[1]{{\leavevmode \color{red} #1}}

\maketitle

\begin{abstract}

Terrain traversability in unstructured off-road autonomy has traditionally relied on semantic classification, resource-intensive dynamics models, or purely geometry-based methods to predict vehicle-terrain interactions.
While inconsequential at low speeds, uneven terrain subjects our full-scale system to safety-critical challenges at operating speeds of 7--10 \si{m/s}.
This study focuses particularly on uneven terrain such as hills, banks, and ditches. These common high-risk geometries are capable of disabling the vehicle and causing severe passenger injuries if poorly traversed.
We introduce a physics-based framework for identifying traversability constraints on terrain dynamics. Using this framework, we derive two fundamental constraints, each with a focus on mitigating rollover and ditch-crossing failures while being fully parallelizable in the sample-based Model Predictive Control (MPC) framework. In addition, we present the design of our planning and control system, which implements our parallelized constraints in MPC and utilizes a low-level controller to meet the demands of our aggressive driving without prior information about the environment and its dynamics.
Through real-world experimentation and traversal of hills and ditches,
we demonstrate that our approach captures fundamental elements of safe and aggressive autonomy over uneven terrain.
Our approach improves upon geometry-based methods by completing comprehensive off-road courses up to 22\% faster while maintaining safe operation.

\end{abstract}

\newcommand{\R}{\ensuremath{\mathbb{R}}}\
\newcommand{\indic}[1]{\ensuremath{\mathbf{1}\left\{#1\right\}}}
\renewcommand{\vec}[1]{\ensuremath{\mathbf{#1}}}
\newcommand{\xb}{\mathbf{x}}
\newcommand{\ub}{\mathbf{u}}
\newcommand{\NN}{\mathcal{N}}
\newcommand{\diag}{\mathrm{diag}}
\newcommand{\one}{\mathds{1}}
\newcommand{\fdiff}[1]{\ensuremath{\frac{\prescript{\mathcal{#1}}{}{d}}{dt}}}
\newcommand{\diffdt}{\ensuremath{\frac{d}{dt}}}
\newcommand{\deltadt}{\ensuremath{\frac{\Delta}{\Delta t}}}
\newcommand{\idiff}{\fdiff{I}}
\newcommand\numberthis{\addtocounter{equation}{1}\tag{\theequation}}
\newcommand{\norm}[1]{\ensuremath{\left\lVert #1 \right\rVert}}
\newcommand{\abs}[1]{\ensuremath{\left\lvert #1 \right\rvert}}
\newcommand{\moi}[1]{\ensuremath{\mathbb{I}_{#1}}}
\newcommand{\angvel}{\ensuremath{\boldsymbol{\omega}}}
\newcommand{\angacc}{\ensuremath{\boldsymbol{\alpha}}}
\newcommand{\acc}{\ensuremath{\vec{a}}}
\newcommand{\roll}{\ensuremath{\phi}}
\newcommand{\pitch}{\ensuremath{\theta}}
\newcommand{\yaw}{\ensuremath{\psi}}
\newcommand{\pos}[2]{\ensuremath{\vec{r}_{#1/#2}}}
\newcommand{\restorque}[1]{\ensuremath{\boldsymbol{\tau}^{\text{res}}_{#1}}}
\newcommand{\restorquep}{\restorque{P}}
\newcommand{\refframe}[2]{\ensuremath{{\mathcal{#1}_{#2}}}}
\newcommand{\bframe}[1]{\ensuremath{\refframe{B}{#1}}}
\newcommand{\bfp}{\bframe{P}}
\newcommand{\comscalar}[3]{\ensuremath{ {#1#2_{#3}} }}
\newcommand{\comg}[1]{\ensuremath{ {#1} }}
\newcommand{\comgp}{\comg{P}}
\newcommand{\resditch}{\ensuremath{\tau_{\text{ditch}}^{\text{res}}}}
\newcommand{\rr}{\ensuremath{{RR}}}
\newcommand{\rrmax}{\ensuremath{\rr_{\text{max}}}}

\newcommand{\cost}[1]{\ensuremath{c_{\text{#1}}(\xb_t, \ub_t)}}

\newbool{short}
\booltrue{short}

\newcommand\blfootnote[1]{%
  \begingroup
  \renewcommand\thefootnote{}\footnote{#1}%
  \addtocounter{footnote}{-1}%
  \endgroup
}

\renewcommand\sout{\bgroup\markoverwith{\textcolor{red}{\rule[0.5ex]{2pt}{0.4pt}}}\ULon}

\section{Introduction}
\begin{figure}[t]
    \includegraphics[width=\linewidth]{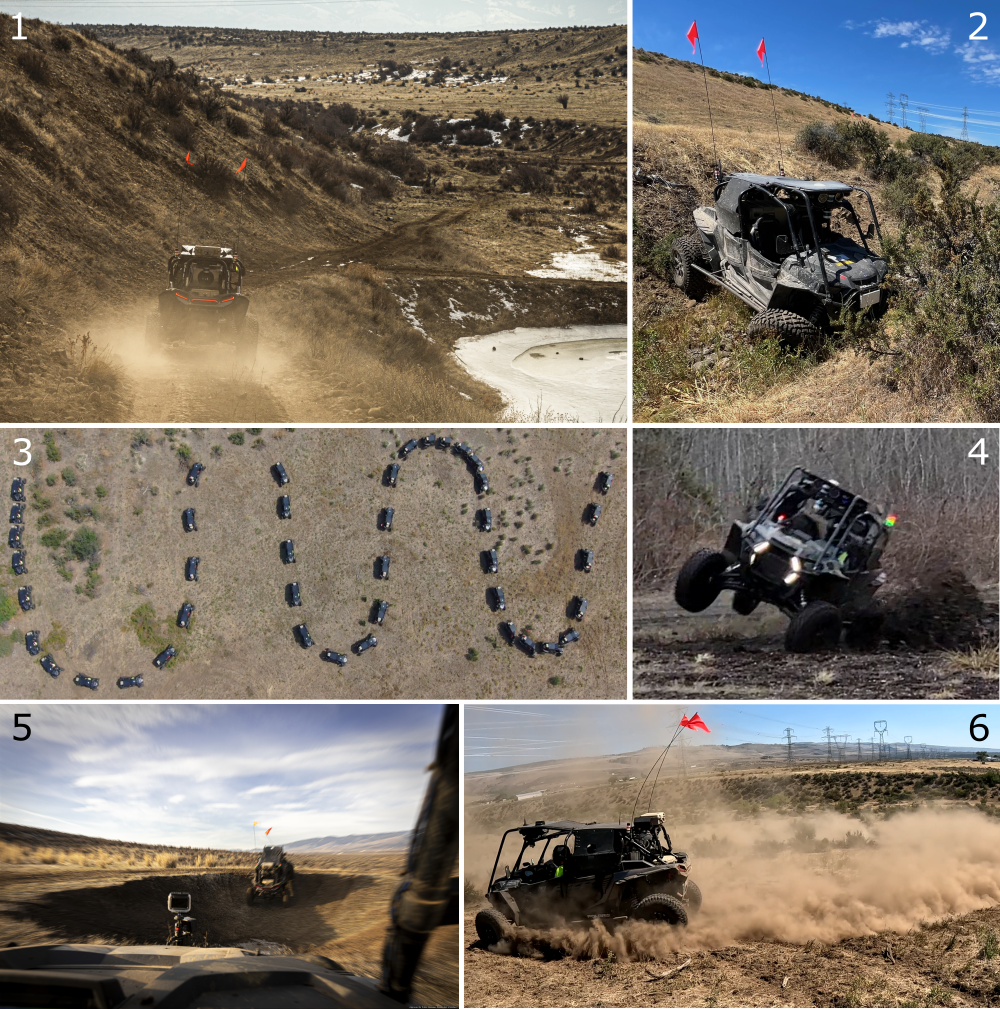}
    \caption{
    (1) Robot in uneven terrain.
    (2) Poor ditch traversal.
    (3) Slalom on a steep hill.
    (4) Poor rollover handling.
    (5) Chasing robot through crater ditch.
    (6) Expert driver performing aggressive circles. \textit{Note that (2) and (4) are not outcomes of our approach.}
    \vspace{-12pt}}
    \label{fig:overview}
\end{figure}
Autonomous vehicles have the potential to be applied to agriculture, transportation in underdeveloped regions, and search and rescue \cite{liu2013robotic, mousazadeh2013technical, carruth2022challenges}. To succeed in these applications, autonomous vehicles must be able to operate without a priori knowledge in \emph{unstructured} and \emph{off-road} environments, which often weaken or entirely violate assumptions commonly made in on-road autonomous vehicle research \cite{ma2015onroad, paden2016onroad}.
In particular, unlike the on-road setting, successful off-road driving depends critically on safe traversal of \emph{uneven terrain} such as hills, ditches, banks, and bumps. 
To demonstrate realistic autonomy in unstructured environments, terrain mapping and control optimization are performed onboard and in real-time. No prior information is provided to the system. In this work, we aim to address two significant failure risks encountered by a full-scale off-road autonomous vehicle (Polaris RZR Turbo S4) traveling at speeds of $7$--$10$ \si{m/s} over uneven terrain. Either rollovers or ditches are easily capable of disabling the vehicle and causing severe bodily harm to passengers.

First, \emph{rollover} can occur when the vehicle becomes disabled by tipping onto its sides. The risk of rollovers is substantially higher in off-road driving where steep hills, evasive sharp turns, and vehicles with high centers of mass due to tall suspension systems are common. 
The second failure risk is associated with \emph{ditches}. 
Often characterized by extreme impact forces, snapped or irreparably bent control arms (\cref{fig:overview}.2) are common failures due to poor ditch traversal.

Aggressive navigation through uneven terrain is further complicated by systems considerations such as imperfect perception, uncertainty, and real-time control.
For instance, ground plane estimation is made difficult by the presence of obstacles (tall grass, short bushes and trees, rocks, etc.) in the scene \cite{meng2023terrainnet, maturana2018real, Guan_He_Song_Manocha_Zhang_2022, Castro_Triest_Wang_Gregory_Sanchez_Rogers_Scherer_2023}, which obfuscates the true ground location. As a result, predicted ground planes can often be noisy, inaccurate, and oversmoothed. Still, the downstream controller must be able to tune away the bias introduced by inaccurate but fast perception.
Most importantly, at nominal speeds of $7$--$10$ \si{m/s}, fast planning and optimization times are safety critical. Low control rates cause the system to plan through misrepresented or delayed semantic obstacles and terrain geometries which can be extremely dangerous. Evidently, the high-speed regime requires that perception and control be connected intimately to ensure safe operation. In addition to challenging sub-systems, high-speed off-road driving is a highly complex systems problem. To the authors' best knowledge, no prior work has demonstrated high speeds in uneven terrain on a full-scale system. We refer readers to our careful review of current literature across related off-road, on-road, small-scale, and low-speed systems in \cref{sec:related}.

We present two main contributions to off-road autonomy in this paper. 
First, we propose a physics-based framework to derive traversability constraints for a mobile robot on uneven terrain. 
We show that these constraints are time-independent and can be parallelized across both samples and the control horizon, enabling aggressive control through a \textit{practical compromise between simple geometry-based methods and inertial dynamics models.}
Second, we report our complete planning and control design with heuristics for off-road navigation validated on a full-scale Polaris RZR S4 1000 equipped with onboard sensing and compute. The planning and control system consists of a planner based on the model predictive path integral (MPPI) algorithm \cite{wagener2019online, williams2017information}, and a low-level feedforward-feedback controller.
Our method is qualitatively validated by expert driver behavior and traverses up to 22\% faster than the geometry-based baseline on two and three kilometer comprehensive off-road courses with vegetation, steep hills, and sharp ditches. 

\section{Related Work}\label{sec:related}

Currently, there are few works which conduct autonomous off-road experimentation on a full-scale platform, fewer which demonstrate speeds beyond 7 \si{m/s}, and, to the authors' best knowledge, none which additionally deploy in real, uneven terrain.
However, work on the same chassis, sensor, and compute platform does exist \cite{gibson2023, Frey_Khattak_Patel_Atha_Nubert_Padgett_Hutter_Spieler_2024, cmuracer}.
These works focus primarily on individual sub-systems, validate with offline data and methods, and do not report on system performance or speeds.
Our work presents the most complete description and evaluation of a control system for the fastest full-scale, off-road, autonomous vehicle in uneven terrain to date. For more distant but related work, we note that due to lack of standardization in the off-road setting, difficulty of the environment and expected performance can vary drastically. Prior works on real, full-scale vehicles do not attempt to navigate in uneven terrain~\cite{Coombs_Murphy_Lacaze_Legowik_2000, Stavens_Hoffmann_Thrun_2007}. However, even in mild terrain conditions, our operating speeds of 7–10 \si{m/s} are higher or comparable to other work.
In~\cite{triest2023learning}, the system navigates autonomously at about 3.2 \si{m/s} and averages 4 interventions over 1.6 km. In the DARPA Grand Challenge, the fastest method achieves speeds between 9–11 \si{m/s} on largely flat terrain, slowing to 4-6 \si{m/s} for mild bumps \cite{Stavens_Hoffmann_Thrun_2007}. Older work also demonstrate capabilities up to 9.7 \si{m/s} over flat ground and binary traversability~\cite{Coombs_Murphy_Lacaze_Legowik_2000}.

Past works and systems reveal that high-speed navigation itself is not a new challenge.
It is the combination of high speeds and challenging, uneven terrain which current methods struggle with.
Our real testing environment is most similar to the environment \textit{simulated} by Lee et al. \cite{Lee_Kim_Mun_Lee_2023} where the proposed MPPI-based approach also averages 7-11 \si{m/s} in simulation. Importantly, the authors empirically discover that a cost must be applied to vertical accelerations and torques in order to maintain control. In the DARPA Subterranean Challenge, Fan et al. constrain velocity in the kinodynamic planner without rigorous justification but state that these constraints reduce energy of interactions with the environment \cite{fan2021step}. 
Evidently, there is a need for systems to require constraints due to the challenging terrain geometry of these unstructured environments. However, no principled justification has yet been provided. We aim to justify these heuristics such that they are motivated through physical modeling and constraints.

Existing work on real systems on uneven terrain but at low speeds generally assume the vehicle adheres to the ground at all times and constrain attitude angles as estimated by only the ground geometry~\cite{datar2023learning,fan2021step,krusi2017driving}. This method is relatively fast, simple to implement, and effective at low speeds \cite{howardkelly}. However, at higher speeds, the method becomes increasingly inaccurate and unsafe due to dynamic effects from unmodeled forces and vehicle inertia.
On the other hand, complex dynamics models can explictly predict wheel forces and trajectories but are often too slow for safe operation. 
For instance, \cite{yu2021nmpc} implements a 3D rigid-body model with tire forces in MPC at a control rate of 3 Hz, about one tenth of our target rate. In \cite{liu2022rmpc}, the control rate is sufficient at 50 Hz but operates at speeds of 3 \si{m/s} over gentler and differentiable ground manifolds. 
Note that due to the architecture of common off-road perception systems, elevation maps are often outputs of complex neural networks \cite{meng2023terrainnet, Frey_Khattak_Patel_Atha_Nubert_Padgett_Hutter_Spieler_2024} or approximations to point clouds \cite{krusi2017driving}. Neither representation admits a practical path toward real-time optimization using aforementioned methods.

Some works have pursued learning-based approaches to driving through uneven terrain. In~\cite{cesar2022towards}, the authors propose using imitation learning for safe traversal of various types of ditches at 1 \si{m/s} in simulation. However, success rates drop by as much as 50\% depending on which types of train and test geometries are selected. Learning-based dynamics modeling has also been proposed in the literature~\cite{gibson2023,triest2022tartandrive,xiao2021learning}. However, in the aggressive regime, collecting data near the safety limits of the vehicle can be dangerous and infeasible in practice. In~\cite{Maheshwari_piaug}, the authors seek to generalize low speed training data to higher speed tests using physics-based modeling, improving tight turns on flat ground over a basic bicycle model.

In summary, current methods fall short in our setting for three main reasons: (1) implementations on real systems lack models of dynamic effects, constraining autonomy to low speeds or mild terrain; (2) full vehicle dynamics models with explicit wheel forces are too computationally expensive; and (3) state-of-the-art real-time perception representations do not lend themselves to otherwise sufficiently fast gradient-based optimization.
In this work, we propose contending with dynamic effects by estimating the implicit forces of sampled paths, similar in principle to methods in~\cite{li2019, Fork_Tseng_Borrelli_2022, howardkelly}.

\section{Physics-Based Traversability Constraints}
\label{sec:cost}
 
\begin{figure*}[ht]
    \centering
    \includegraphics[width=.55\linewidth]{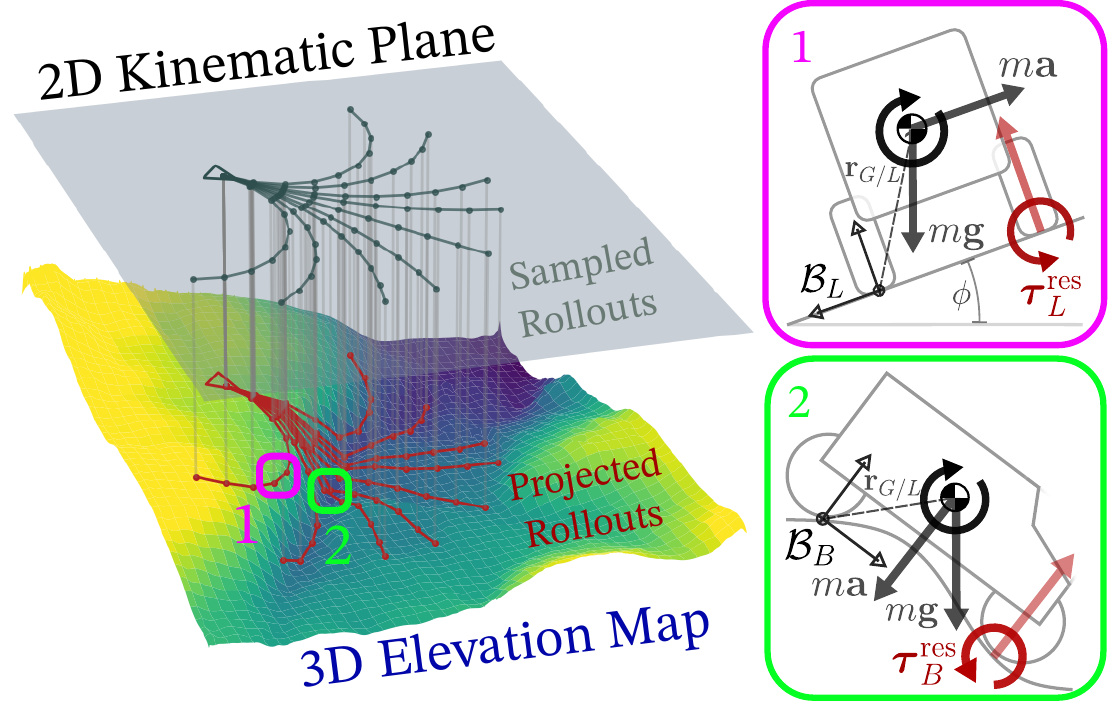}
    \caption{Rollouts are sampled in the 2D kinematic plane then constrained by their implicit dynamics on the elevation map (left). Frames and forces for rollover analysis (rear-view, upper-right) and for ditch analysis (side-view, bottom-right).}
    \label{fig:traversability}
    \hfill\newline
    \includegraphics[width=.5\linewidth]{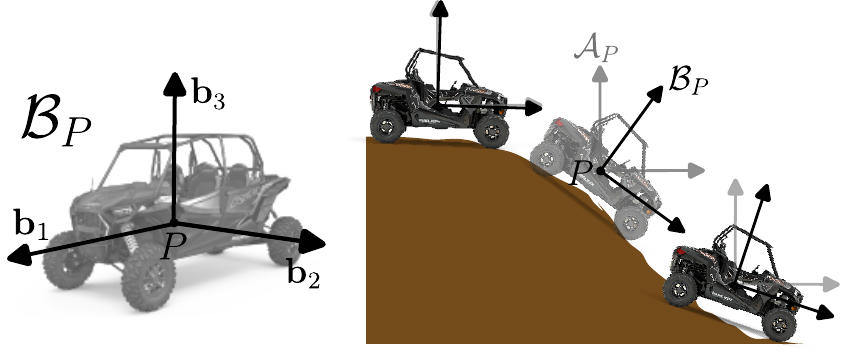}
    \includegraphics[width=.17\linewidth]{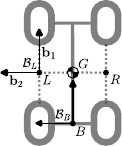}
    \caption{Illustrations of body frame \bframe{P} on vehicle (left), dynamics of \bframe{P} and path frame \refframe{A}{P} (middle), and analysis frames \bframe{B} and \bframe{L} (right). }
    \label{fig:body-frame}
\end{figure*}

In automobile safety testing, metrics such as static stability factor~\cite{huston2014another,pai2017trends}, lateral load transfer ratio~\cite{doumiati2009tire,larish2013pltr}, and time to rollover~\cite{chen2001differential,dahmani2013vehicle} quantify the risk of vehicle rollovers by focusing on the inertial forces which may tip the vehicle. 
This line of work supports the intuition that the ground and path geometry alone do not adequately represent the vehicle dynamics. However, these geometries, when coupled with gravitational forces and vehicle inertia, should contain sufficient information about feasible traversal speeds.
In this section, we derive constraints on the vehicle by incorporating geometrical information in a physics model, which is used to inform cost function design for MPPI in~\cref{sec:system}. 

\subsection{Dynamics Model with Residual Torque}

Following our rationale, our modeling objective will be to derive a relationship between the geometry of the environment and the relevant forces. As a motivating thought experiment, consider a marble rolling down a track shaped like a downwards-facing parabola. Guided by gravity, the marble will adhere to the track without issue. Once the marble is pushed, it begins to lose its grip on the track. Pushed harder still, the marble will begin to lift away from the track until its trajectory is entirely separated. For our vehicle, we are concerned with \textit{any} wheel leaving the ground rather than the entire vehicle all at once. Thus, we are interested in constraining the rotational rate of the vehicle which is a tighter constraint on the state space. We proceed with the three-dimensional, rigid-body analogue of this thought experiment. 

We conduct analysis with respect to an intermediate non-rotating frame $\refframe{A}{P}$ with globally-fixed basis vectors and origin $P$ fixed to some point on the vehicle. Vectors are demarcated by bold symbols as opposed to unbolded scalars. The $\times$ operator is the cross product. Let $\pos{G}{P}$ be the position of the vehicle's center-of-mass $G$ relative to $P$.
The relevant torques on the vehicle include: gravity ($\pos{G}{P} \times m\vec g$, with $m$ the mass of the vehicle), inertial torque (simplified as $\pos{G}{P} \times \acc$), and torques due to wheel forces $\boldsymbol \tau_P$. This leads us to the following torque-balance model,
\begin{align}
    \moi{P} \cdot \boldsymbol{\alpha} &= \vec r_{G/P} \times m \left(\vec g - \acc \right) + \restorquep, \label{restorque}
\end{align}
where $\angacc$ is the angular acceleration and $\moi P$ is the moment of inertia tensor about $P$. 
What is $\restorquep$? In the following derivations, we will choose $P$ such that the frictional (lateral) wheel forces have negligible torque contribution about $P$. As a result, $\boldsymbol{\tau}_P$ becomes directly proportional to the normal forces at the wheels. As it is also technically possible for the path dynamics to demand \textit{negative} normal forces to satisfy the torque balance, we instead use the fictional ``residual'' torque $\restorquep$.
As we proceed, we vary our reference frame origin $P$ (e.g. $P\coloneqq L$ or $P\coloneqq B$) depending on the failure mode of interest.

\subsection{Rollover} \label{sec:rollover}

On-road, rollovers generally occur when a turn is made so sharply that large lateral forces are induced as a result of vehicle inertia. Without sufficient downward force to counteract this lateral or tipping force, this results in vehicle rollover.
Rollover prevention for automobiles is well-studied and methods in essence constrain the normal forces at the vehicle's wheels \cite{larish2013pltr, chen2001differential, tota2022alg, Fork_Tseng_Borrelli_2022, howardkelly}. 

Let $\mathcal{B}_{P}$ represent the body frame with origin $P$ which rigidly rotates with the vehicle. Let the basis unit vectors be $\vec b_1$ pointed forward in the direction of vehicle motion, $\vec b_2$ pointed to the left, and $\vec b_3$ pointed upwards away from the ground, completing the right-handed frame.
Without loss of generality, we begin with modeling left rollover so that we are interested in the rotational forces which might cause the vehicle to roll counterclockwise along the $\vec b_1$ direction. Let $L$ lay on the line from the front left to rear left wheels while aligning laterally with the center of mass such that $\pos{G}{L} \cdot \vec b_1 = 0$.
Now using the body frame $\mathcal{B}_L$ for left rollover analysis, we want to bound $\restorque{L}$ such that the wheel forces are always acting in support of the vehicle. This suggests that $\restorque{L} \cdot \vec b_1 < 0$ (see Fig. \ref{fig:traversability} for diagram), i.e.,
\begin{align*}
    \left[ \moi{L} \cdot \angacc - \pos{G}{L} \times m \left(\vec g - \acc \right) \right] \cdot \vec b_1 &< 0 \numberthis \label{l_rollover}.
\end{align*} 
Observe that if these constraints were violated such that $\restorque{L} \cdot \vec b_1 \geq 0$, our torque-balance would require the wheel forces to \textit{pull} the vehicle down or, more realistically, result in \textbf{wheel airtime}. Likewise for right rollovers, the analysis is similarly conducted about $R$ on the opposite side of the vehicle but with a change of sign such that $\restorque{R} \cdot \vec b_1 > 0$.

\subsubsection*{2D Case} %

We consider the non-accelerating two-dimensional case and make the simplifying assumption that our rolling angular acceleration is small, $\angacc \cdot \vec b_1 \approx 0$. Let $P$ be either $L$ or $R$ such that $\vec{r}_{G/P} \cdot \vec b_1 = 0$ (see Fig. \ref{fig:body-frame}).

Note we can also decompose the acceleration $\vec a$ by differentiating the vehicle velocity $\vec{v} \coloneqq v \vec b_1$,
\begin{align*}
    \vec{a} &= \diffdt \vec{v} = \diffdt [v] \vec b_1 + v \diffdt [\vec b_1] \\
    &= \dot{v} \vec b_1 + v \angvel \times \vec b_1 \numberthis \label{accel_decomp}
\end{align*}\label{eq:acc_decomp}
where we have used the fact that $\diffdt [\vec b_1] = \angvel \times \vec b_1$ \cite{kasdin2011}. 
The residual torque from \eqref{l_rollover} can be combined with the decomposition from \eqref{accel_decomp} and our non-acceleration assumption such that
\begin{align*}
    \restorquep \cdot \vec b_1 &= \left[ \moi{P} \cdot \angacc - \vec r_{G/P} \times m \left(\vec g - \acc \right) \right] \cdot \vec b_1 \\
    &=  - \left[ \vec r_{G/P} \times m \left(\vec g - \acc \right) \right] \cdot \vec b_1 \\
    &=  - \left[ \vec r_{G/P} \times m  (\vec g - \dot{v} \vec b_1 - v \angvel \times \vec b_1 )\right] \cdot \vec b_1 \numberthis \label{2d_rollover1}
\end{align*}
We note the vector product identity $(\vec a \times \vec b) \cdot \vec c = (\vec a \times \vec c) \cdot \vec b$ which allows us to further simplify such that
\begin{align*}
    (\vec r_{G/P} \times \dot{v} \vec b_1) \cdot \vec b_1 =  ( \dot{v}\vec b_1 \times \vec b_1) \cdot \vec r_{G/P} = 0
\end{align*}
A final vector triple product identity $(\vec a \times \vec b) \times \vec c = (\vec a \cdot \vec c) \vec b -  (\vec a \cdot \vec b) \vec c$  allows further algebraic simplification,
\begin{align*}
    \vec r_{G/P} \times (\angvel \times \vec b_1) &= (\vec r_{G/P} \cdot \vec b_1) \angvel - (\vec r_{G/P} \cdot \angvel) \vec b_1\\
    &= - (\vec r_{G/P} \cdot \angvel) \vec b_1
\end{align*}
where we have used the fact that $\vec r_{G/P} \cdot \vec b_1 = 0$ .

Reintroducing our simplifications into \eqref{restorque} gives
\begin{align}
  \restorque{P} \cdot \vec b_1 = -m  (\pos{G}{P} \times \vec g) \cdot \vec b_1 - mv (\vec r_{G/P} \cdot \angvel). \label{rollover_vector}
\end{align} 

We can further simplify and scalarize by rewriting our basis vectors using known state quantities. Let $\roll$ be the roll angle of the vehicle such that
$\vec g = g\sin\roll \vec b_2 - g\cos\roll \vec b_3$
with gravitational constant $g$. Let the position of the center of gravity in the body frame be given by 
$\vec r_{G/P} = \comgp_{2} \vec b_2 + \comgp_{3} \vec b_3$
. For simplicity, we approximate the angular velocity here to be such that $\angvel \approx \omega_3 \vec b_3 \approx v \kappa \vec b_3$ where $\kappa$ is the steering curvature 
Then substituting our expressions for $\vec g$, $\angvel$, and $\vec r_{G/P}$ into \eqref{rollover_vector} gives,
\begin{align*}
  \restorquep \cdot \vec b_1  = - [\comgp_{2}g\cos\roll + \comgp_{3} (v^2 \kappa + g\sin\roll)]
\end{align*}
Depending on whether $P \coloneqq L$ or $P \coloneqq R$, the sign of $\comgp_{2}$ and direction of inequality in \eqref{l_rollover} changes (fig. \ref{fig:traversability}). Both constraints can be simplified as
\begin{equation}
    \abs{\comgp_3 (v^2 \kappa + g\sin\roll)} < {\comgp_2 g\cos\roll} \label{2d_rollover_bound},
\end{equation}
which gives our physics-based rollover constraint in terms of our kinematic state and elevation map angles.

\subsection{Ditches} %

\label{sec:ditches}

In traversing ditches, vehicle-ground separation almost surely leads to extreme impact forces as the vehicle wheels are reintroduced to the ground plane. Thus, like rollover prevention, we aim to discourage this separation in the first place.
Our approach to the ditch constraint is similar to rollover prevention but instead focused on the front-back (about $\vec b_2$) rotational forces. Let $B$ be between the rear left and right wheels, aligned with the center of mass such that $\pos{G}{B} \cdot \vec b_2 = 0$. The residual torque $\restorque{B}$ should again always act in support of the vehicle such that $\restorque{B} \cdot \vec b_2 < 0$ (see Fig. \ref{fig:traversability}). However, we find that this sole constraint is insufficient for ditch handling. In fact, it is also necessary to impose that the residual torque is not overly excessive. Namely, $\restorque{B} \cdot \vec b_2 > \tau_{\text{min}}^{\text{res}}$ where $\tau_{\text{min}}^{\text{res}}$ is a safety parameter. Observe that $\restorque{B}$ is directly related to the normal force at the front wheels. Thus, $\tau_{\text{min}}^{\text{res}}$ corresponds to the greatest allowable normal force at the front wheels that we are willing to incur for the sake of mechanical stresses or rider comfort. We can summarize our constraint with the following bound:
\begin{align}
    \tau^{\text{res}}_{\text{min}} < \restorque{B} \cdot \vec b_2 < 0. \label{eq:ditch_bound}
\end{align}
\subsubsection*{2D Case}

We derive a form for this inequality that is computable under our MPPI framework in the two-dimensional case.
Again using the path frame decomposition from \eqref{accel_decomp} and~\cite{kasdin2011}, the residual torque about $B$ can be written as
\begin{align*}
    \restorque{B} &= \moi{B} \cdot \angacc - \vec r_{G/B} \times m \left(\vec g - \dot{v}\vec b_1 - v\angvel \times \vec b_1 \right).
\end{align*}
In frame $\bframe{B}$, we have $\vec r_{G/B} = \comg{B}_1 \vec b_1 + \comg{B}_3 \vec b_3$. We consider a strictly pitching maneuver such that $\angvel \approx \omega_2 \vec b_2$ and $\vec g = g\sin\pitch \vec b_1 - g\cos\pitch
\vec b_3$ where $\pitch$ is the pitch of the vehicle. Substituting these terms,
\begin{small}
\begin{align*}
    \restorque{B}\cdot \vec b_2 &= \bar I_{22} \alpha_2 + \comg{B}_{1} (v\omega_2 + g\cos\pitch\numberthis) +  \comg{B}_3 (\dot v + g\sin\pitch),
\end{align*} \label{eq:specific}
\end{small}
where $\bar I_{22}$ is the specific product of inertia about $\vec b_2$. For conciseness, define $\tau^{\text{res}}_{\text{ditch}} \coloneqq \restorque{B}\cdot \vec b_2$ such that reintroducing \eqref{eq:ditch_bound} gives us our simplified bound in terms of our kinematic state, elevation map, and allowable torque parameter $\tau^{\text{res}}_{\text{min}}$,
\begin{equation}
    \tau^{\text{res}}_{\text{min}} < \tau^{\text{res}}_{\text{ditch}} < 0. \label{eq:2d_ditch_bound}
\end{equation}

In summary, we have derived values for both rollover prevention and ditch handling that are strictly functions of our sampled state $\xb = \begin{pmatrix}
    x & y & \yaw
\end{pmatrix}$ and controls $\vec u = \begin{pmatrix}
    v & \kappa
\end{pmatrix}$ and whose constraint computations are simple box bounds. Importantly, these constraints can be computed in-place, are time-independent, and are parallelizable across rollout sampling \textit{and} the control horizon.  This is critical for the fast and efficient computation of these constraints on our full system whose details we provide in the next section. A theoretical and empirical analysis is also made to better determine the errors introduced from our simplifying assumptions, which can be found in \cref{sec:constraint_anal}.

\section{System Overview}\label{sec:system}

Our system can be abstracted into an upstream perception model, mid-level sample-based MPC planner (MPPI), and a low-level controller (fig. \ref{fig:architecture}). Our robotic platform is a modified Polaris RZR S4 1000, fitted with LiDAR and RGB-D cameras as well as four NVIDIA GeForce RTX 3080 GPUs. Note the control algorithm utilizes only one of these GPUs.

\subsection{Perception}

The perception model receives data from the onboard LiDARs and cameras as input. At a rate of about 10 Hz, it outputs two grid-maps with a radius of 50 meters centered at the vehicle's current location. One of these grid-maps, the costmap, provides semantic information about the environment (e.g. trees, grass, rocks)~\cite{chitta2021, kelly2006, shaban2022bevnet}. The second grid-map, the elevation map, provides elevation predictions at each corresponding location in the map. This is a common representation for ground geometry where sensor input is provided by LiDAR or stereo cameras \cite{fankhauser2018, forkel2022, stolzle2022, triebel2006}.
As done in \cite{yu2021nmpc, gibson2023, liu2022rmpc}, we use these elevations to calculate predicted attitude angles (roll $\roll$ and pitch $\pitch$) as demonstrated in Fig. \ref{fig:traversability}. These angles are used to inform our autonomy in case of potential violations to our constraints.

\subsection{MPPI}

\subsubsection{Preliminaries}
Let $\xb_t$ and $\ub_t$ represent state and control at time $t$. For a given control horizon $H$, we randomly sample $N$ open-loop control sequences $\{\ub_{t:t+H-1}^i\}_{i=1}^N$. For each sequence, we compute their cumulative costs over the horizon
$C(\ub_{t:t+H-1}^i;\xb_t) \coloneqq \sum^{H-1}_{h=0} c(\xb_{t+h}, \ub_{t+h})$
using the predicted trajectory rollout given by the bicycle kinematics model~\cite{kong2015bicycle}. Without a priori knowledge of the environment, a model of the terrain dynamics can be impractical or altogether inaccessible. As inaccurate global models are known to result in poor performance for MPC \cite{williams2017information}, the abstracted bicycle model leverages the lower-level PI controller to better handle unmodeled terrain dynamics. 
With this model, our state can be represented by $\xb_t = \begin{pmatrix}
    x_t & y_t & \yaw_t
\end{pmatrix}$
where $\begin{pmatrix}
x_t & y_t
\end{pmatrix}$ is the position of the center of mass and $\yaw_t$ is the heading. Our control is then represented by $\ub_t = \begin{pmatrix}
    v_t & \kappa_t
\end{pmatrix}$ where $v_t$ is the linear velocity and $\kappa_t$ is the steering curvature.
To obtain the optimal control command for the vehicle, we use the update law given by the model-predictive path integral (MPPI) control algorithm~\cite{williams2017model,williams2017information, bhardwaj2022storm}. This family of MPC algorithms have become increasingly popular due to their flexibility with non-differentiable cost functions and parallelizability,
especially with a kinematic model \cite{bhardwaj2022storm, williams2017information}.

\begin{figure}[t]
    \centering
    \includegraphics[width=0.9\linewidth]{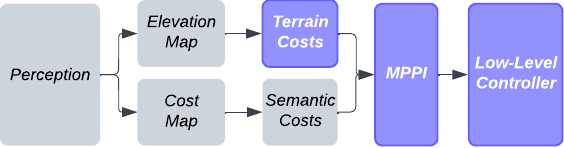}
    \caption{Information flow across perception and control. The focus of this work are highlighted in purple.
    }
    \label{fig:architecture}
\end{figure}

\subsubsection{Fast and Parallel Model Rollouts}

We implement MPPI sampling, kinematic rollout, and cost function evaluation in a TorchScript PyTorch module running on an NVIDIA RTX 3080 GPU. Cost computations are vectorized across samples and the control horizon such that one iteration of MPPI samples are evaluated altogether in parallel. Our average MPPI computational time for one iteration is $29.6\pm0.9$ ms or about $33.8$ Hz.

\subsubsection{Cost Function} \label{cost-terms}
We evaluate costs of sampled kinematic rollouts against the elevation map. Our cost function \footnote{The full cost function includes more costs for handling the costmap, goal progress, etc. which we provide detail in appendix \ref{sec:other-costs}} is given by:
\begin{equation*}\small
    c(\vec x, \vec u) \coloneqq w_1c_{\text{rollover}}(\vec x, \vec u) +\, w_2 c_{\text{airtime}}(\vec x, \vec u) +\, w_3 c_{\text{bump}}(\vec x, \vec u) 
\end{equation*}
where each $w_i$ is a hand-tuned weight, typically a large positive value to help disqualify unsafe trajectories from the control optimization. 

Incorporating our rollover bound \eqref{2d_rollover_bound}, the rollover cost term is a cumulative sum over the constraint violations,
\begin{equation}\small
    c_{\text{rollover}}(\vec x_{t+h}, \vec u_{t+h}) \coloneqq \sum_{k=t}^{t+h} \rr(k) \cdot \indic{\rr(k) > \rrmax}
\end{equation}
where,
\begin{equation}
    \rr(k) \coloneqq \frac{\abs{v_k^2 \kappa_k - g\sin\roll(k)}}{\cos\roll(k)} \label{eq:rr}
\end{equation}
and $\rrmax \leq g\,\comgp_2/ \comgp_3$ is the tunable parameter to control the vehicle's aggressiveness with rollover risk. 

For ditch handling, in addition to the maximum allowable torque magnitude $\tau_{\text{min}}^{\text{res}}$, we introduce a tunable upper bound $\tau_{\text{max}}^{\text{res}}$ such that $\tau^{\text{res}}_{\text{ditch}} < \tau_{\text{max}}^{\text{res}} < 0$. While in theory, $\tau_{\text{max}}^{\text{res}}$ can be 0, this corresponds to the vehicle ``hovering'' above the ground, which can result in loss of control or instability.
As the linear accelerations $\dot v_t$ sampled from MPPI can be noisy, 
we remove $\dot v_t$ from \eqref{eq:2d_ditch_bound} to obtain a more tunable bound. We also compute $\omega_2$ and $\alpha_2$ through forward numerical differencing of $\pitch(t)$. Our time-indexed $\resditch$ can be computed as
\begin{small}
\begin{align}
    \resditch(t) = \bar I_{22} \alpha_2(t)
    + \comgp_{1} v_t \omega_2(t) - \comgp_{3} g\sin\pitch(t) - \comgp_{1} g\cos\pitch(t) \nonumber
\end{align}
\end{small}
For conciseness, define $\delta_{\text{min}}(t) \coloneqq \tau_{\text{min}}^{\text{res}} - \resditch(t)$ and $\delta_{\text{max}}(t) \coloneqq \resditch(t) - \tau_{\text{max}}^{\text{res}}$. As with rollover, we employ the cumulative sum of violations,
\begin{small}
\begin{align}
    c_{\text{airtime}} (\vec x_{t+h}, \vec u_{t+h}) &\coloneqq \sum_{k=t}^{t+h} \delta_{\text{max}}(k) \cdot \indic{\delta_{\text{max}}(k) > 0} \label{eq:airtime} \\
    c_{\text{bump}}(\vec x_{t+h}, \vec u_{t+h}) &\coloneqq \sum_{k=t}^{t+h} \delta_{\text{min}}(k) \cdot \indic{\delta_{\text{min}}(k) > 0} \label{eq:bump}
\end{align}
\end{small}

The cumulative (running) sum in \eqref{2d_rollover_bound} and \eqref{eq:2d_ditch_bound} deserve special attention. This heuristic is important because violations of our constraint almost surely imply violations of our modeling assumptions later in the horizon. For instance, if the vehicle is predicted to leave the ground, subsequent predictions have higher uncertainty and should be penalized accordingly. The cumulative sum also strongly penalizes sooner violations over later violations in the horizon.

\subsubsection{Tuning}

We find rollover parameter $\rrmax$ is theoretically and empirically correlated with lateral force on the vehicle. To tune this parameter, we begin at $\rrmax\coloneqq 2$ and increase the value to $\rrmax\coloneqq 3.4$ where the human safety operator deems the behavior too risky. This is a qualitative assessment due to the diversity of terrain conditions but can generally include occurrences such as slipping, tipping, or instability.

We find ditch parameter $\tau_{\text{max}}^{\text{res}}$ correlates with reductions while $\tau_{\text{min}}^{\text{res}}$ correlates with increases in normal force as perceived by the operator. Starting with $\tau_{\text{max}}^{\text{res}}\coloneqq -5$ and $\tau_{\text{min}}^{\text{res}}\coloneqq -7$, this bound is widened to $\tau_{\text{max}}^{\text{res}}\coloneqq -3$ and $\tau_{\text{min}}^{\text{res}}\coloneqq -10$ where the safety operator deems the behavior too risky or intolerable. If the operator feels the vehicle lifting away from the ground too quickly, $\tau_{\text{max}}^{\text{res}}$ is lowered. If the operator deems ground impact too forceful, $\tau_{\text{min}}^{\text{res}}$ is raised.

\subsubsection{Handling Control Delays}
In practice, the issued control command $\hat{\ub}_t$ can not be immediately executed on the vehicle due to software and hardware delays. This means that we actually have $\xb_{t+1} = f(\xb_t, \hat{\ub}_{t-\tau})$ for some delay $\tau\geq 0$. To handle delays, we consider the MPC optimization problem of horizon $H$ from a projected future state $\Tilde{\xb}_t$ (rather than the true state $\xb_t$). The projected future state $\Tilde{\xb}_t$ is given by rolling out the previously issued controls $(\hat{\ub}_{t-\tau},\ldots,\hat{\ub}_{t-1})$ starting from state $\xb_t$.
These prepended controls effectively simulate the control commands that are issued but yet to be executed on the system. 

\begin{figure*}[t]
    \centering
    \includegraphics[width=.9\linewidth]{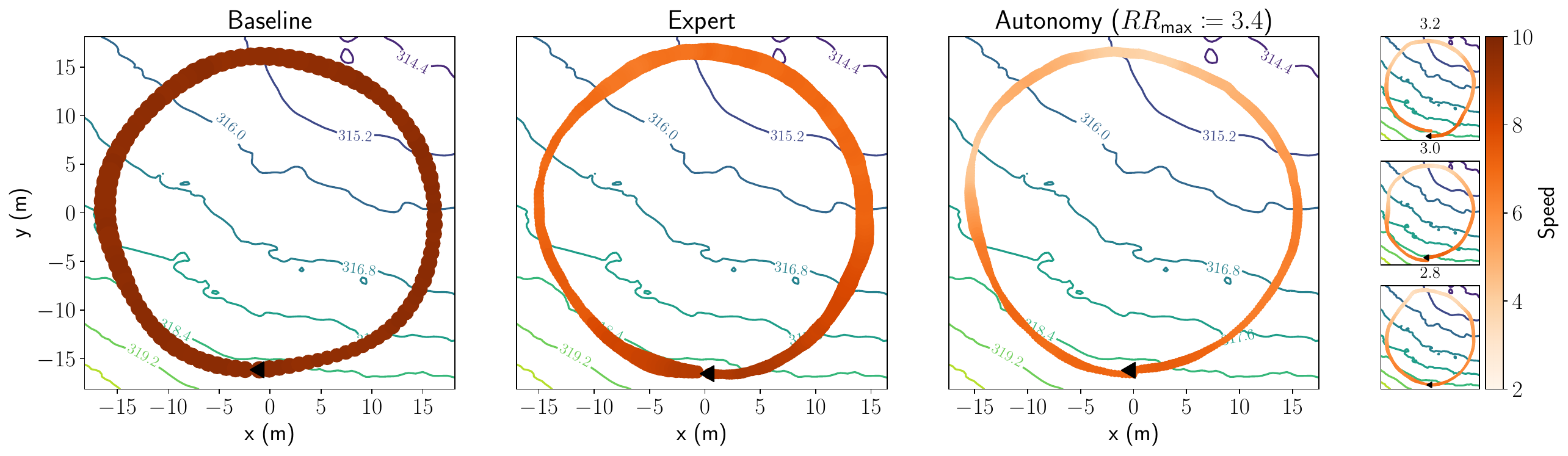}
    \includegraphics[width=\linewidth]{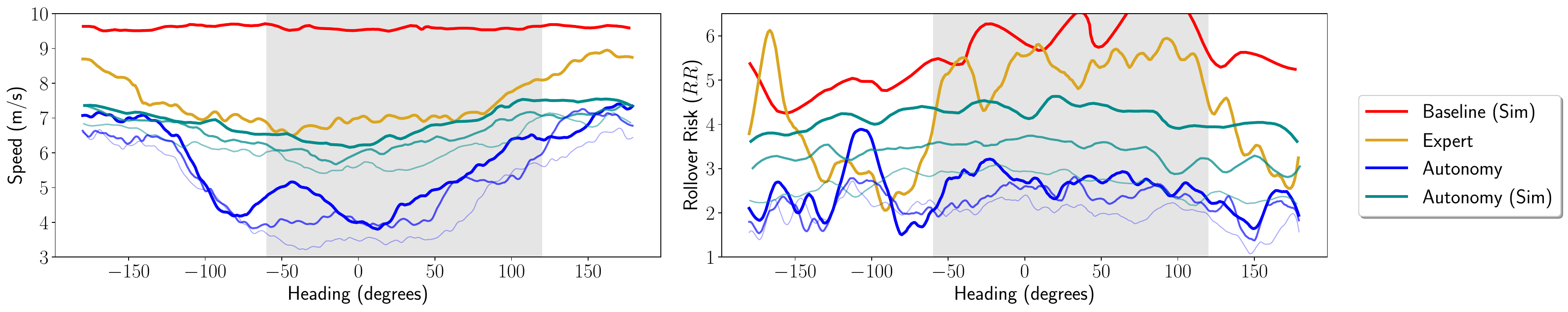}
    \caption{\textit{Top row}. Spatial plot over elevation contour map of 15-meter radius circles on a 10-degree incline hill. Results for each experiment are averaged over at least five loops. Speed and $RR$ are reflected by the line color and marker width, respectively. Autonomy results of decreasing $\rrmax$ control parameter are shown in the right column. \textit{Bottom row}. Plots of speed and rollover risk as a function of heading angle. Off-camber turning highlighted by grey box. Thicker and opaquer autonomy lines correspond with higher $\rrmax$ settings (Autonomy $\rrmax\in\{3.0, 3.2, \textbf{3.4}\}$; Autonomy (Sim) $\rrmax\in\{4, 5, \textbf{6}\}$). Autonomy (Sim) simulates MPPI for higher than tolerable $\rrmax$ values. Baseline results are simulated due to unsafe behavior. } 
    \label{fig:expert-auto-sim-rr}
\end{figure*}

\subsubsection{Sampling Control Sequences} \label{sec:sampling-control}

The original MPPI algorithm samples the control sequences $\{\ub_{t:t+H-1}^i\}_{i=1}^N$ from a Gaussian distribution centered around a nominal control sequence given by the shift operator \cite{williams2017model,williams2017information, bhardwaj2022storm, wagener2019online}. In our implementation, we sample raw control sequences $\{\Tilde{\ub}_{t:t+H-1}^i\}_{i=1}^N$ from a mixture of Gaussian distributions\footnote{See appendix \ref{sec:mixture-gaussian} for more details on the mixture distributions used.}, and process the raw control sequences to generate sampled control sequences. When necessary, these Gaussians help the optimization with: (1) denser sampling around the nominal trajectory, (2) abrupt slow-downs, and (3) resetting local convergence on high-cost regions.

\paragraph{Sampling feasible control sequences}

The sampled changes in linear velocity and curvature must be limited due to the vehicle's throttle, brake, and steering limitations. Additionally due to static friction, the steering is unchanged at linear velocities near zero.

We process the raw control sequences $\{\tilde\ub_{t:t+H-1}^i\}_{i=1}^N$ so that feasibility constraints are satisfied. Let $\tilde\ub_{t+h}^i=(\tilde v_{t+h}^i, \tilde\kappa_{t+h}^{i})$, we iteratively compute the processed control sequence $\ub_{t+h}^i=( v_{t+h}^i, \kappa_{t+h}^{i})$, 
\begin{equation}\small
    \begin{split}
        v_{t+h}^i &= \mathrm{clip}(\tilde{v}_{t+h}^i, v_{t+h-1}^i\pm\Delta v_{\max}),\\
        \kappa_{t+h}^{i} &= \left\{
        \begin{array}{ll}
            \kappa_{t+h-1}^{i} & \text{if }|v_{t+h}^i| < v_{\min},\\
            \mathrm{clip}(\tilde{\kappa}_{t+h}^{i}, \kappa_{t+h-1}^{i}\pm\Delta\kappa_{\max}) & \text{otherwise},
        \end{array}\right.
    \end{split}
\end{equation}
where $\mathrm{clip}(a, b\pm c)\coloneqq {\max}({\min}(a, b+c), b-c)$ for $a,b,c\in\R$ and $c>0$. $\Delta v_{\max}$ and $\Delta\kappa_{\max}$ are the maximum changes of linear velocity and curvature, respectively, per time step, and $v_{\min}$ is the minimum speed that allows change of steering.
We take $(v_{t-1}^i, \kappa_{t-1}^{i})$ as the control issued at the previous step, $\hat{\ub}_{t-1}$, for all $i$.

\subsection{Low-Level Controller}
\label{sec:controller}

Off-road environments have wildly varying ground conditions (friction, compliance, looseness, roughness, etc.) which can make direct optimization over throttle extremely challenging as this requires a second-order model over the environment. As done in the DARPA Subterranean and DARPA Grand Challenge \cite{Hoffmann_Tomlin_Montemerlo_Thrun_2007, Thakker_Alatur_Fan_Tordesillas_Paton_Otsu_Toupet_Agha-mohammadi_2021}, we depend upon a low-latency controller to mitigate these unmodeled dynamics.
We employ a low-level PI feedback controller to attain desired speeds from MPPI using vehicle throttle and brake:
\begin{equation*}
\label{eq:low-level-control}
u = -K_P(v-v_t) - K_I\int(v-v_t)dt + c_1 g\sin\pitch + c_2 v
\end{equation*}
where $v_t$ is the desired linear velocity from MPPI, and $v$ is the actual velocity of the vehicle. The feedforward terms $c_1 g\sin\pitch + c_2 v$ compensate for the acceleration in the $\vec b_1$ direction from gravity and the acceleration due to drag and friction. We determine the linear coefficients, $c_1$ and $c_2$ empirically from data and tune the proportional and integral gains $K_P$ and $K_I$. $u$ is then converted to throttle (when $u$ is positive) or brake (when $u$ is negative) using a pre-identified fixed mapping. Finally, for the steering angle, we directly translate the desired curvature $\kappa_t$ from MPPI to the steering angle command with a pre-identified fixed mapping.

\section{Results}\label{sec:results}

Our experiments consist of two groups of isolated tasks and three (two plus one reverse) complex courses. The first isolated task involves completing circles on slopes and is meant to test the
robot's
ability to avoid rollovers. The second task is crossing various ditches.
We evaluate the geometry-based method (the baseline)\footnote{for details about the geometry-based method see appendix \ref{sec:baseline-impl}}, the physics-based method (ours; also referred to as the autonomy), and human-driven results (the expert). We refer readers to the paper's 
website
for videos of our experiments and general off-road performance.

\begin{figure*}[t]
    \centering
    \includegraphics[width=.89\linewidth, trim={0 0cm 0 0.5cm}, clip]{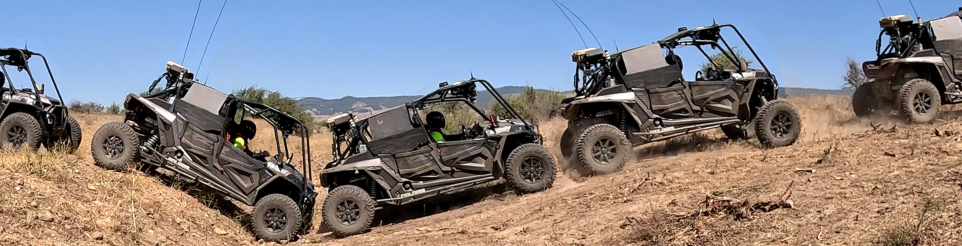}\vspace{.3cm}
    \includegraphics[width=.9\linewidth, trim={2cm 0 0 0}]{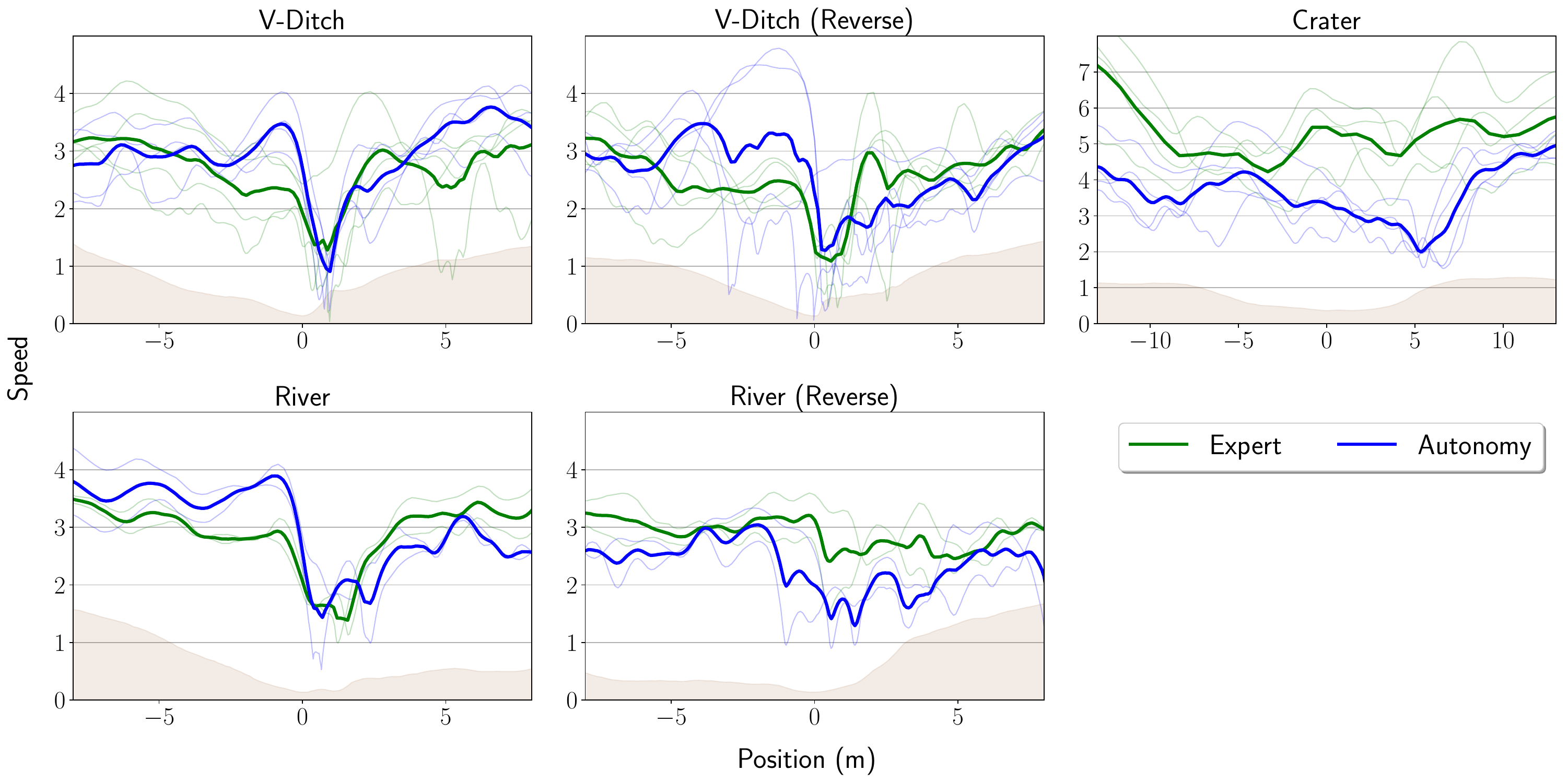}
    \caption{Robotic platform crossing the V-Ditch (pictured above; driver is present only for safety purposes). Average velocity is plotted as a function of the position through the ditch over the elevation profile (brown). Individual ditch traversals are plotted transparently.}
    \label{fig:vditch}
\end{figure*}

\subsection{Rollover}
We consider a baseline algorithm which uses a cost that penalizes roll angles above $20$ degrees, similar to~\cite{datar2023learning,fan2021step,krusi2017driving}.
We compare the expert to the baseline in simulation (Fig. \ref{fig:expert-auto-sim-rr}).
While the expert consistently operates at about $\rr(t) \leq 6$, the baseline goes well above this threshold and in theory can surpass this arbitrarily as the slope angle never exceeds $20$ degrees. At $\rrmax \approx 6$, the expert is already experiencing loss of control from slipping and tilting. Far exceeding this threshold, the baseline would likely cause a rollover or severe loss of control in the real world.

We also compare the expert to our method, with parameter $\rrmax \coloneqq 3.4$ (Fig. \ref{fig:expert-auto-sim-rr}). For safety reasons, we do not autonomously operate the real vehicle at $\rrmax > 3.4$. 
Instead, we conduct real and simulated tests at 6 different settings of $\rrmax$ to show that autonomous behavior indeed trends safely toward expert behavior with increasing values of $\rrmax$. In particular, the expert drives fastest through the on-camber turn and slowest through the off-camber turn (also observable in the slalom in Fig. \ref{fig:overview}.3). Note that MPPI often operates below its $\rrmax$ setting due to the penalizing tails of the control Gaussian \cite{williams2017information}. The expert also likely drives faster during the on-camber turn due to suspension loading which is not modeled in our constraints. Nonetheless, the autonomy consistently and safely remains below expert speeds with some tunable margin as desired.

\subsection{Ditches}
Similar to the rollover task, we seek to qualitatively evaluate our method in ditch crossing by comparing it with the expert (Fig. \ref{fig:vditch}). Without explicit velocity constraints, this experiment verifies that the vehicle adequately traverses through ditches in a safe manner. The tests are conducted over three distinct ditches with varying geometries. In these experiments, the autonomy's maximum speed was limited to 5 \si{m/s} for safety.

We note a couple differences between the Autonomy and Expert that can be attributed to the Expert's access to prior knowledge about the environment.
The autonomy's slower crossing speeds in the River (reverse) and Crater ditches are largely due to the sharp exiting ridges. Proceeding at the expert-executed speeds would result in infringing on the ditch upper-bound $\tau^{\text{res}}_{\text{max}}$ and excessively reducing contact with the ground. As the expert has prior knowledge about what exists beyond the ridge, they are more comfortable with momentarily reducing control authority in return for speed.
Secondly, the expert decelerates much earlier than the autonomy before the crossing. Note that the descent angle of the ditch is often occluded from the vehicle's LiDAR and camera during the approach and can result in delayed sensing of the ditch geometry. This results in the autonomy's sharp decelerations in the V-Ditch and river crossings whose descent angles are initially occluded but revealed last-minute. However, as the expert has knowledge of the occluded regions of the ditch, they are able to make more comfortable decelerations in anticipation.

\begin{figure*}[t]
    \centering
    \includegraphics[height=14.3cm]{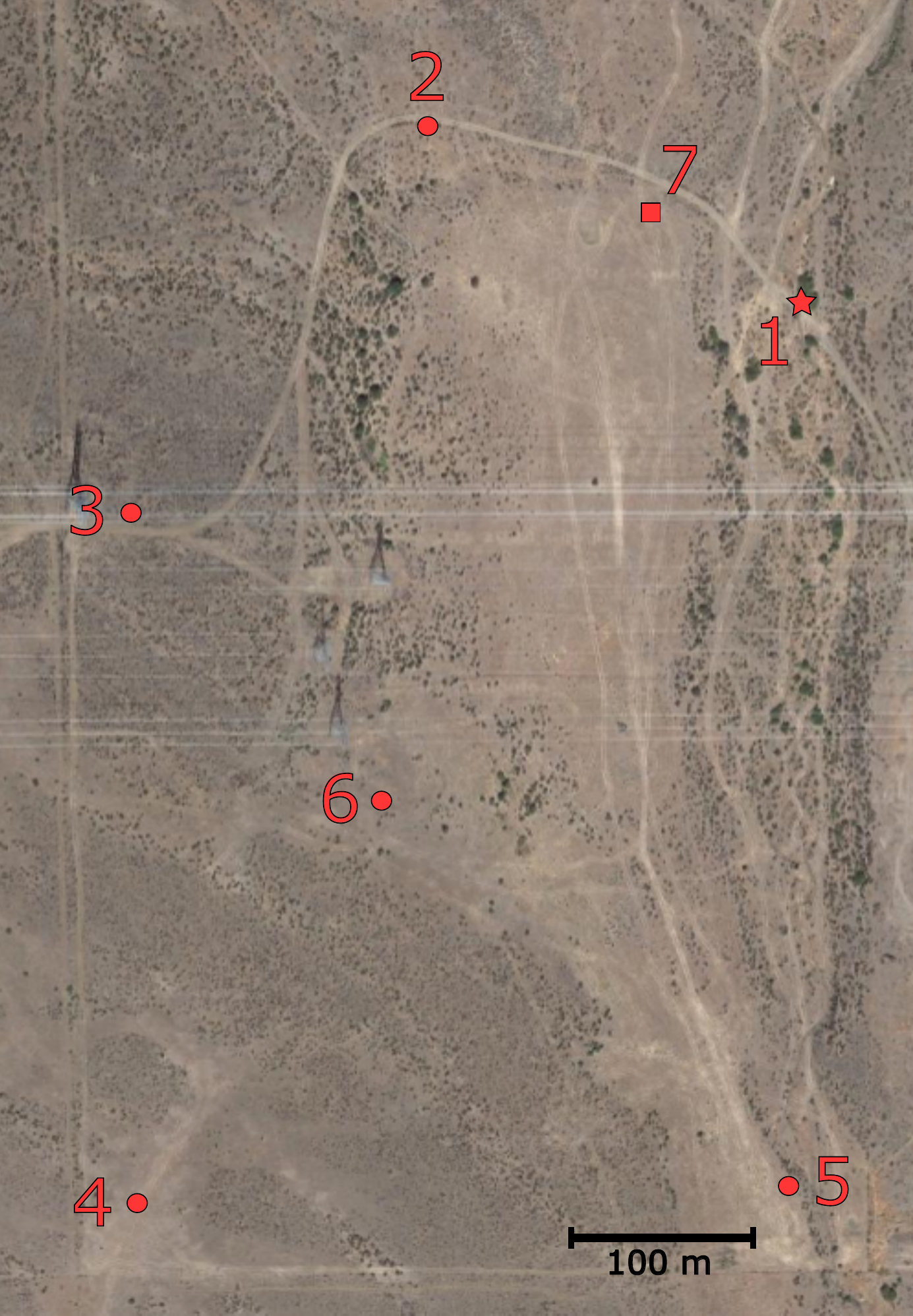}
    \hspace{6pt}
    \includegraphics[height=14.3cm]{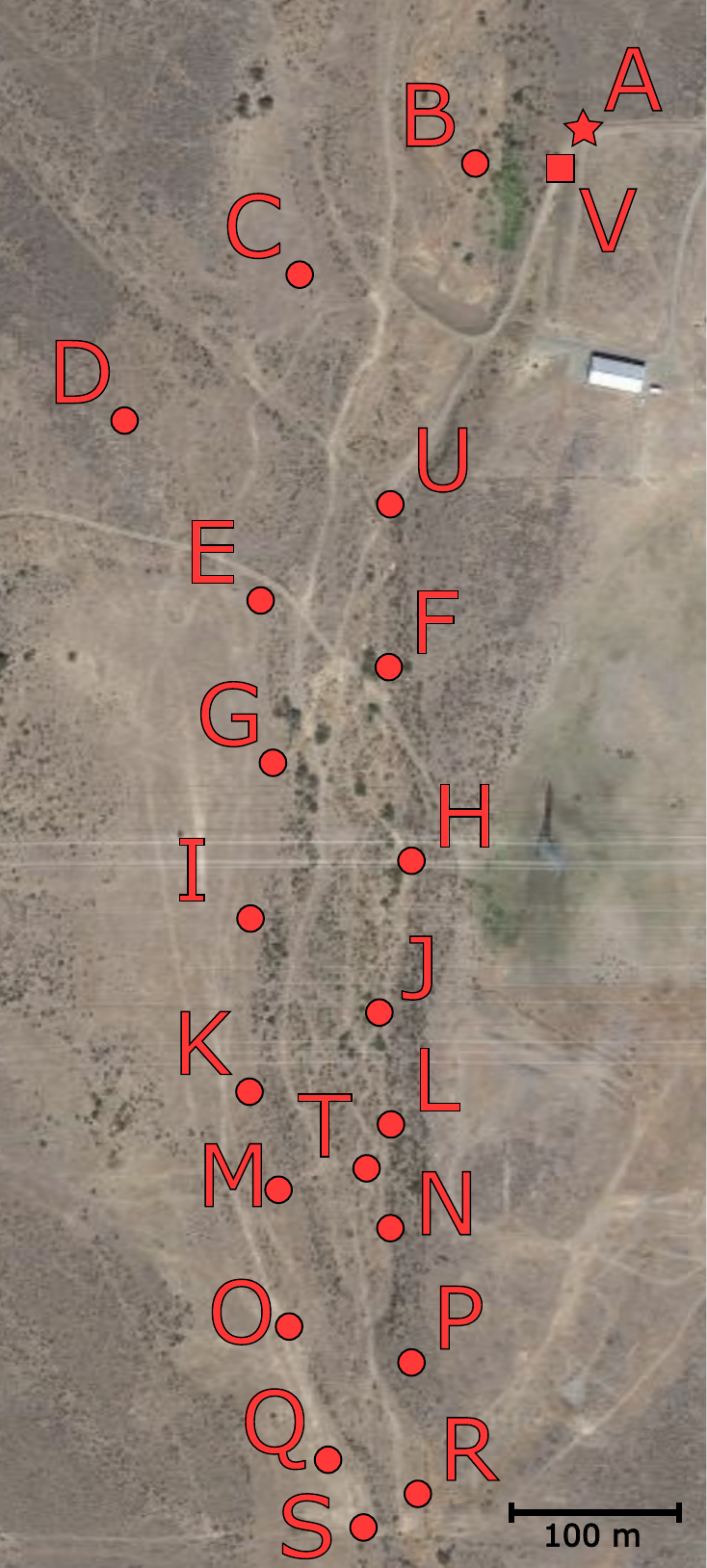}
    \caption{Complex courses Tour (left) and Baja (right) with numbered and alphabetized waypoints. The Tour is approximately two kilometers in total distance with $\pm 80$ \si{m} of elevation change. Reaching waypoints \textit{4} through \textit{7} require traversal of steep slopes and ditches. Waypoints \textit{1} through \textit{4} test trail-following and nominal behaviors (obstacle avoidance and basic navigation and control). Baja is approximately three kilometers and repeatedly crosses the region densest in ditches and vegetation from waypoints \textit{E} through \textit{U}.}
    \label{fig:tour}
\end{figure*}

\subsection{Complex Course}

Our ultimate objective is to autonomously navigate through complex terrain. To evaluate this, we run the autonomy through the Tour, the Ruot (backwards Tour), and the Baja (Fig. \ref{fig:tour}). To evaluate method performance, we are interested in three key metrics:

\subsubsection{Interventions}

An intervention occurs when the human safety operator (on standby during autonomous operation) must make a manual maneuver (e.g. quickly stop or evade) to reset the vehicle due to poor or dangerous decision-making. More interventions generally correlate with ``worse'' autonomy.

\subsubsection{Course Time}

The course time is the total time it takes the vehicle to complete the course by reaching every waypoint in order. An aim of aggressive driving is to (safely) reduce travelling time.

\subsubsection{Ditch False-Positives}

A falsely identified ditch (false-positive or FP) is a common complication in the development of a ditch control method due in part to imperfect perception. Because safe ditch traversal speeds are typically around 1-3 \si{m/s}, unnecessary slowdowns from our nominal speeds of 7-10 \si{m/s} are detrimental to aggressive autonomy.

We find that the geometry-based baseline is more susceptible to ditch FPs despite best efforts in parameter tuning to maintain safe ditch traversal while reducing FPs. It is likely due to pure geometrical reasoning, which treats all detected ditches equally regardless of current speed or actual severity. Moreover, there is less flexibility in the tunable parameter space to compensate for this, since we have shown in section \ref{sec:cost} the severity is necessarily correlated with traversal speeds. The accumulation of multiple FPs inevitably increases the course time on the Tour for the baseline. With our method on the Ruot, 
we flag a potential FP encountered when cresting a steep hill (waypoint \textit{5} to \textit{4}). Without slippage modeling, MPPI believes its command speed will result in airtime \eqref{eq:airtime} at the crest. In reality, the vehicle lacks momentum and traction to cleanly crest the hill and begins to stall. An intervention is made to prevent damage to the transmission belt. On the Baja, which involves primarily ditches and obstacle avoidance, false-positives are less likely due to the dense presence of ditches. However, some of these ditches do not require full slowdowns for safe traversal. The baseline method does not accommodate this and results in longer course completion times due to over-slowing.
\\

\begin{table}
    \begin{tabular}{lrrrrrrr}\toprule
        &\multicolumn{3}{c}{\textbf{Tour}}&\multicolumn{2}{c}{\textbf{Ruot}}&\multicolumn{2}{c}{\textbf{Baja}}
        \\\cmidrule(r){2-4}\cmidrule(r){5-6}\cmidrule(r){7-8}   
        & None & Base & Ours & Base & Ours & Base & Ours\\\midrule
        Course Time & 5:26    & 6:01 & 5:28
                & 5:35 & 5:48
                & 11:22 & 8:54 \\
        Ditch FP   & N/A & 8 & 0
                & 6 & 1
                & 0 & 0 \\
        Interventions  & 1   & 0 & 0 
                & 0 & 1
                & 0 & 0
        \\\bottomrule
    \end{tabular}
    \caption{Run statistics on complex courses (see Fig. \ref{fig:tour}). Base indicates the geometry-based baseline. Our method is the physics-based approach in this work. ``None'' indicates no ditch-control method as a part of the autonomy. }\label{tab:complex-course}
\end{table}

\section{Conclusion}

In this study, we propose a framework to augment 2D kinematic sampling and real-time elevation maps with physics-based constraints for safe and aggressive off-road autonomous driving. This framework yields two constraints addressing failures due to rollover and unsafe ditch handling which are fully parallelizable for sample-based MPC. Our control system design integrates this approach with a fast and efficient implementation of the MPPI algorithm. Finally, we validate and compare our method to a purely geometry-based baseline and an expert driver through real-world experiments. These experiments include autonomously traversing hills, ditches, and full complex courses on our full-scale robotic platform.

We identify a few limitations to our methods which we should be pursued in future work.
While kinematic sampling is fast and parallelizable, uneven terrain with increasingly high-frequency features (e.g. rock beds) demand denser sampling methods to provide adequate signal for our constraints to behave as expected.
Secondly, the impact force of a ditch cannot sometimes be strictly modeled as a path-induced normal force, i.e. collisions. Assuming such an environment can be properly perceived, predicting collisions like these at present can be computationally prohibitive.
Our operating environments can also range widely with no prior information and our current methods can require significant tuning. Adaptive or learning-based methods for dynamics models for control can be highly impactful for improving system-wide performance in this setting.
Finally, by comparing our method's results to an expert driver, we observe similarities that demonstrate our approach's ability to capture fundamental elements of safe yet aggressive control. However, beyond physical capabilities, the expert's ability to reason about uncertainty sets it apart from the autonomy. Future work in this area can be hugely impactful in the real-world setting.

\section*{Acknowledgment}
This research was developed with funding from the Defense
Advanced Research Projects Agency (DARPA). The views, opinions and/or findings expressed are those of the author and should not be interpreted as representing the official views or policies of the Department of Defense or the U.S. Government. This material is also partially based upon work supported by the National Science Foundation Graduate Research Fellowship under Grant No. DGE-2140004. Any opinions, findings, and conclusions or recommendations expressed in this material are those of the author(s) and do not necessarily reflect the views of the National Science Foundation.

\newpage
\bibliographystyle{plainnat}
\bibliography{ref}

~\\
\twocolumn[
\centerline{\fontsize{18}{18}\selectfont \textbf{
        Supplementary Materials
}}
~\\
]
\renewcommand{\arraystretch}{1.25}
\section{Appendix}
\begin{table}[t]
    \centering
    \begin{tabular}{l l l r}
    \toprule
    & & \multicolumn{2}{c}{\textit{Maximum Constraint Error}} \\
    \cmidrule(l){3-4}
        Quantity & Assumption & Theoretical & Empirical \\
    \midrule
        \multirow{3}{*}{$\rrmax$} & $\angacc \cdot \vec b_1 \approx 0$ &
        $\bar{I}_{11} \abs{\alpha_1}$
        & 0.01 \\
        & $\angvel \approx \omega_3 \vec b_3$ &
        $\abs{v} \norm{\vec r_{G/P}}  \norm{\angvel - \omega_3 \vec b_3}$
        & 0.9 \\
        & $\omega_3 \approx v\kappa$ &
        $\abs{P_3 v} \abs{\omega_3 - v\kappa}$
        & 0.8 \\
        \midrule
        \multirow{1}{*}{$\tau_{\text{min}}^{\text{res}},\,\tau_{\text{max}}^{\text{res}}$} & $\angvel \approx \omega_2 \vec b_2$ & 
        $\abs{v} \norm{\vec r_{G/B}}  \norm{\angvel - \omega_2 \vec b_2}$
        & 2.8 \\
        \midrule
        \multirow{1}{*}{$v$} & $v_{\text{cmd}} \approx v_{\text{actual}}$ &
        $\abs{v_{\text{cmd}} - v_{\text{actual}}}$
        & 0.15 \\
    \bottomrule
    \end{tabular}
    \caption{Empirical Effects of Simplifying Assumptions on Constraints}
    \label{tab:violations}
\end{table}
\renewcommand{\arraystretch}{1}

\subsection{Explicit Forms of Gaussians for Control Sampling}\label{sec:mixture-gaussian}

\subsubsection{Conventional Samples}

The conventional samples are drawn from a Gaussian distribution centered at the nominal control sequences,
$$\Tilde{\ub}_{t:t+H-1}^\mathrm{conv}\sim \NN(\bar\ub_{t:t+H-1}, I_H \otimes \Sigma),\, \Sigma = \diag([\sigma_1, \sigma_2]).$$

\subsubsection{Narrow Samples}
We find it to beneficial to sample more control sequences that are close to the nominal control sequence. Therefore, we sample a number of control sequences with a smaller covariance,
$$\Tilde{\ub}_{t:t+H-1}^{\mathrm{narrow}}\sim \NN(\bar\ub_{t:t+H-1}, I_H \otimes (s_{n}\cdot\Sigma)),$$ with $s_{n}\in(0,1)$ being the narrow scaling coefficient. 

\subsubsection{Scaled Samples} We additionally sample around a scaled nominal control sequence which has smaller linear velocities. This allows the vehicle to effectively slow down while following a similar path as the nominal trajectory. 
Let $\bar\ub_{t+h}=(\bar v_{t+h}, \bar\kappa_{t+h}^d)$ for any $h\in[0, H-1]$. We define the scaled nominal control sequence as 
$$\bar\ub_{t:t+H-1}^\mathrm{scaled} \coloneqq ((s_s\bar v_t, \bar\kappa_t^d), \ldots, (s_s\bar v_{t+H-1}, \bar\kappa_{t+H-1}^d)),$$
with $s_s\in(0,1)$ being scaling coefficients. The scaled samples are drawn from a Gaussian distribution centered at the scaled nominal control sequence,
$$\Tilde{\ub}_{t:t+H-1}^\mathrm{scaled}\sim \NN(\bar\ub_{t:t+H-1}^\mathrm{scaled}, I_H \otimes \Sigma).$$

\subsubsection{Reset Samples}
Finally, we include a small number of raw control sequences sampled from fixed-mean Gaussian distributions. These samples can reset the optimization if the nominal control sequence becomes high-cost due to, e.g., newly detected obstacles and terrain features.
$$\Tilde{\ub}_{t:t+H-1}^{\textrm{reset},k}\sim \NN(\one_H \otimes \bar\ub_\mathrm{reset}^k, I_H \otimes \Sigma).$$
In particular, we consider three fixed mean controls, $\ub_\mathrm{reset}^1 = (0,0)$, $\ub_\mathrm{reset}^2 = (0, -\kappa^d_{\max})$ and $\ub_\mathrm{reset}^3 = (0, \kappa^d_{\max})$, where $\kappa_{\max}^d$ is the maximum desired curvature. These fixed mean controls all uses zero linear velocity but have different desired curvatures.

\subsection{System Parameters and Other Costs}
\label{sec:other-costs}

Here, we include our settings for the primary design parameters of MPPI and a brief explanation of the other primary costs used in the system.

\subsubsection{Goal Costs}
In full course runs, the `goal' point for MPPI is dynamically set by a higher level planner as the true goal points can be many kilometers apart. We have often found it useful to reconsider the mathematical measure of distance between two points. In short, we employ a time-to-go metric using the distance of the dubins path between points and our predicted speed.
\begin{equation*}
    \cost{goal} \coloneqq w_{\text{goal}} \cdot d_\text{dubins}(\xb_t,\xb_{\text{goal}}) / v_t
\end{equation*}
    
\subsubsection{Coplanarity Costs}
The fidelity of our elevation maps result in high uncertainty for areas with high frequency features (loose rock beds, sharp discontinuities, etc.). In addition to helping to ensure our analysis in section \ref{sec:cost} remains valid, We encourage the vehicle to navigate toward low frequency regions by applying cost if the predicted wheel locations are non-coplanar. Let $\xb_\text{fl}$, $\xb_\text{fr}$, $\xb_\text{bl}$, $\xb_\text{br}$ be the four wheel locations and $A = \begin{pmatrix}
        \xb_\text{fl} & \xb_\text{fr} & \xb_\text{bl} & \xb_\text{br}
    \end{pmatrix}^T$. Let $\vec h = \begin{pmatrix}
        h(\xb_\text{fl}) & h(\xb_\text{fr}) & h(\xb_\text{bl}) & h(\xb_\text{br})
    \end{pmatrix}^T$ where $h(\xb)$ is the elevation at $\xb$. Then, the coplanarity cost is
    \begin{equation*}
        \cost{coplanar} \coloneqq w_{\text{coplanar}} \cdot || \vec h_t - A_tA_t^\dagger \vec h_t||
    \end{equation*}
    
\subsubsection{Costmap Costs}

Semantically classified objects are assigned hand-tuned weights that are simply added to the final cost of the trajectory if any part of the vehicle footprint\footnote{footprint($\xb_t$) is the set of all $\begin{pmatrix} x_t & y_t \end{pmatrix}$ positions at time $t$ that corresponds to some point on the vehicle (e.g. the four wheels, the front bumper, etc.)} occupies the same cell. Finally, the cost is scaled with velocity.
    \begin{equation*}
        \cost{costmap} \coloneqq w_{\text{costmap}}~v_t \cdot \sum_{\xb \in \text{footprint}(\xb_t)} \text{costmap}(\xb)
    \end{equation*}

\begin{table}[t]
    \centering
    \rowcolors{1}{gray!25}{white}
    \begin{tabular}{|c  c|}
        \hline
        \rowcolor{gray!50}
        Parameter & Value \\
        \hline
        Temperature & 1.0 \\
        No. of Samples & 10,000 \\
        Horizon (s) & 5.0 \\
        \hline
        $\sigma_1$ & 4.0 \\
        $\sigma_2$ & 2.0 \\
        $\tau_{\text{min}}^{\text{res}}$ & -10.0 \\
        $\tau_{\text{max}}^{\text{res}}$ & -3.0 \\
        $P_2$  & 0.9 \\
        $P_3$  & 1.3 \\
        $B_1$  & 1.8 \\
        $B_3$  & 1.3 \\
        $\roll_{\text{max}}$  & 20$^\circ$ \\
        $\pitch_{\text{max}}$  & 30$^\circ$ \\
        \hline
    \end{tabular}
    \caption{Table of relevant parameter settings. The temperature, number of samples, and horizon are with respect to the MPPI algorithm as laid out in \cite{williams2017information}.}
    \label{tab:parameters}
\end{table}

\subsection{Baseline Implementation}\label{sec:baseline-impl}

The geometry-based baseline is strictly a function of the spatial $\begin{pmatrix}
    x_t & y_t
\end{pmatrix}$ path of the vehicle and the elevation map. Let $\bar c(\cdot)$ indicate cost functions used for the baseline.

\newcommand{\seq}[1]{\ensuremath{\left\{#1\right\}_{h=0}^{H}}}

\subsubsection{Rollover}
Rollover costs using pure geometrical methods treat the vehicle statically and use the critical rollover angle as the costing threshold. However, in practice, we want to constrain the vehicle far below this threshold, and so it is more practical to simply define a hand-tuned angle limit. Let $\beta(t) \coloneqq |\pitch(t)| - \pitch_{\text{max}}$ and $\gamma(t) \coloneqq |\yaw(t)| - \yaw_{\text{max}}$. We apply the cumulative sum of violations,
    \begin{equation*}
        \bar{c}_{\text{pitch}}(\vec x_{t+h}, \vec u_{t+h}) \coloneqq \sum_{k=t}^{t+h} \beta(k) \cdot \indic{\beta(k) > 0} \label{eq:bl-pitch}
    \end{equation*}
    \begin{equation*}
        \bar{c}_{\text{yaw}}(\vec x_{t+h}, \vec u_{t+h}) \coloneqq \sum_{k=t}^{t+h} \gamma(k) \cdot \indic{\gamma(k) > 0} \label{eq:bl-yaw}
    \end{equation*}
    
\subsubsection{Ditches}
This method inspects the geometry in the rollout for deviations from the current vehicle pitch. If a ditch is detected with some certainty, the max vehicle speed will be set to a pre-defined value. In the development of this method, many other approaches were implemented such as using second derivative values of pitch and detecting sudden low elevations. The following implementation performed the best. We use the pitch value at $t$ to predict the expected elevation of the next time $t+1$ and compare this prediction with the actual elevation as provided by the elevation map $h(\xb_{t+1})$. Namely,
    \begin{align*}
        ds &\coloneqq \sqrt{ (x_{t+1} - x_t)^2 + (y_{t+1} - y_t)^2) } \\
        \delta h_{t+1}(\xb_{t+1}) &\coloneqq \tan{(\pitch_t)}  - \left[h(\xb_{t+1}) - h(\xb_{t+1})\right] / ds
    \end{align*}
    Define the ``ditch value'' at $t$ to be the max of $\delta h_t$ over the footprint of the vehicle and ``masked'' by states with downward pitch. This mask reduces false-positives associated with uphills. Explicitly,
    \begin{equation*}
        V_\text{ditch}(\xb_t) \coloneqq \max_{\xb\in\text{footprint}(\xb_t)}\delta h_t(\xb) \cdot \indic{\pitch_t > 0}
    \end{equation*}
    Finally, if $V_\text{ditch}(\xb_t) > V_\text{max}$ anywhere in the nominal (optimized) rollout, the max sampled vehicle speed (see section \ref{sec:sampling-control}) is set to $v_\text{ditch}$ at time $t$. We found that this sampling constraint performed better in practice for this method over costing, due in part to the false positive frequency.

\subsection{Constraint Model Error Analysis}\label{sec:constraint_anal}

In our constraint model, some simplifying assumptions are made to obtain bounds that can be computed in-place without storing excessively large tensors. We summarize the influence of our assumptions by bounding their errors on \eqref{2d_rollover_bound} and \eqref{eq:2d_ditch_bound}.

Recall the unsimplified residual torque about $P$ is given by:
\begin{align}
    \bar{\boldsymbol{\tau}}_P^{\text{res}} = \bar{\mathbb{I}}_P \cdot \angacc - \vec r_{G/P} \times \left(\vec g - \dot{v}\vec b_1 - v\angvel \times \vec b_1 \right) \label{eq:app_restorque}
\end{align}

where we have divided by the mass of the vehicle as in \eqref{2d_rollover_bound} and \eqref{eq:specific}.

\subsubsection{Rollover}

Recall we take $P\in\{L, R\}$. Our first assumption is negligible rolling acceleration, i.e. $\angacc \cdot \vec b_1 \approx 0$.
The error we expect from this assumption is clear from inspecting the first term of \eqref{eq:app_restorque},
\begin{align}
    \bar{\mathbb{I}}_P \cdot \angacc \cdot \vec b_1 = \bar{I}_{11}\alpha_1 \label{eq:R1} \tag{R1}
\end{align}
where we have used a diagonal moment of inertia for simplicity. Next, we assume that the angular velocity is primarily in $\vec b_3$, i.e. $\angvel \approx \omega_3 \vec b_3$. To see the effect of this assumption, we focus our attention to just the affected triple product term in \eqref{eq:app_restorque},
\begin{align}
    \vec r_{G/P} \times \left(\angvel \times \vec b_1 \right) &= \vec r_{G/P} \times \left[(\angvel - \omega_3 \vec b_3 + \omega_3 \vec b_3) \times \vec b_1 \right] \nonumber \\
    &= \vec r_{G/P} \times \left( \omega_3 \vec b_3 \times \vec b_1 \right) \nonumber \\
    &\qquad + \vec r_{G/P} \times \left[(\angvel - \omega_3 \vec b_3) \times \vec b_1 \right] \nonumber \\
    &\leq \underbrace{ \vec r_{G/P} \times \left( \omega_3 \vec b_3 \times \vec b_1 \right)}_{\text{Constrained by } \rrmax} \nonumber \\
    &\qquad + \underbrace{\norm{\vec r_{G/P}} \norm{\angvel - \omega_3 \vec b_3}}_{\text{Error due to Simplifying Assumption}} \label{eq:rr_ineq}
\end{align}
where the first triple-product term in \eqref{eq:rr_ineq} is part of the quantity constrained in \cref{sec:rollover}. The latter term of \eqref{eq:rr_ineq} gives the theoretical maximum error due to our assumption when multiplied with the remaining term $v$ in \eqref{eq:app_restorque},
\begin{align}
\abs{v} \norm{\vec r_{G/P}}  \norm{\angvel - \omega_3 \vec b_3}\label{eq:R2} \tag{R2}
\end{align}

Finally, we also assume that the bicycle model is accurate in estimating the angular velocity, i.e. $\omega_3 \approx \kappa v$. Beginning from \eqref{rollover_vector},
\begin{align}
    v\,(\vec r_{G/P} \cdot \angvel) &\approx  v\,(\vec r_{G/P} \cdot \omega_3 \vec b_3) \nonumber \\
    &= v \left[\vec r_{G/P} \cdot (v\kappa + \omega_3 - v\kappa) \vec b_3\right] \nonumber \\
    &= P_3 v^2 \kappa + P_3v(\omega_3 - v\kappa) \label{eq:steer_anal}
\end{align}
where again the first term in \eqref{eq:steer_anal} is used in the original constraint \eqref{2d_rollover_bound} and the second term is the error term,
\begin{align}
    \abs{P_3 v}\abs{\omega_3 - v\kappa} \label{eq:R3} \tag{R3}
\end{align}

\subsubsection{Ditches}

The assumption used in the derivation for our ditch constraints is $\angvel \approx \omega_2 \vec b_2$. Thus, the bound is equivalent to \eqref{eq:rr_ineq} except 
with $P\coloneqq B$,
\begin{align}
    \abs{v} \norm{\vec r_{G/B}}  \norm{\angvel - \omega_2 \vec b_2} \label{eq:D1} \tag{D1}
\end{align}

\subsubsection{Velocity Tracking}

The tracking assumption due to employing the low-level controller is simply
\begin{align}
    \abs{v_\text{actual} - v_\text{cmd}} \label{eq:V1}\tag{V1}
\end{align}

\subsubsection{Results}

\begin{figure*}
    \centering
    \includegraphics[width=.45\linewidth]{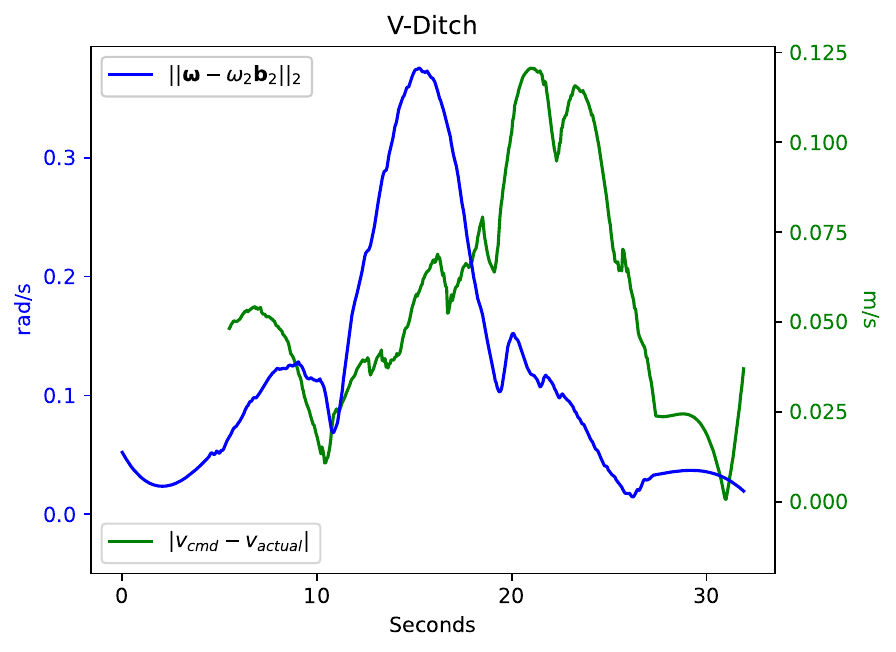}
    \includegraphics[width=.45\linewidth]{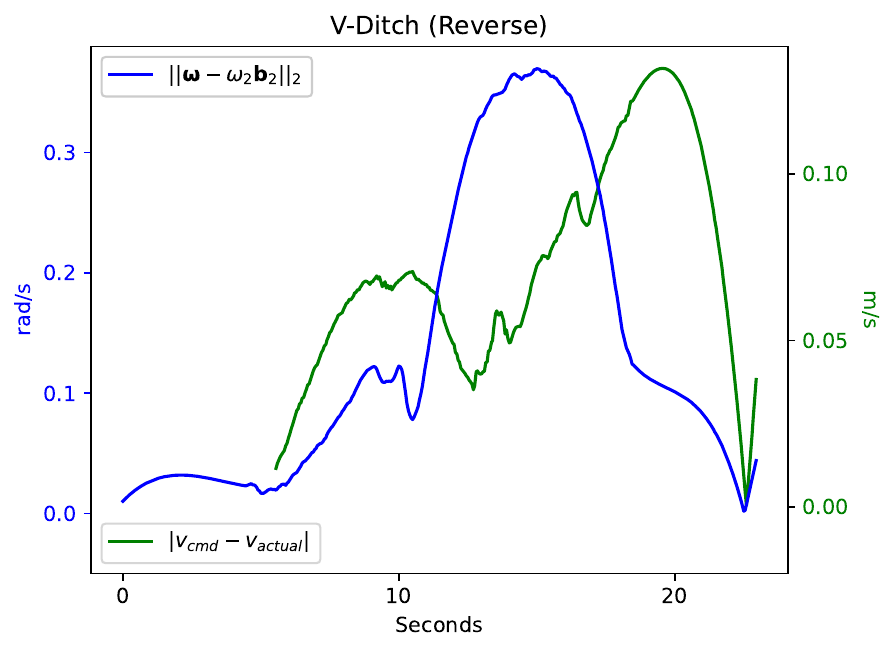}
    \includegraphics[width=.45\linewidth]{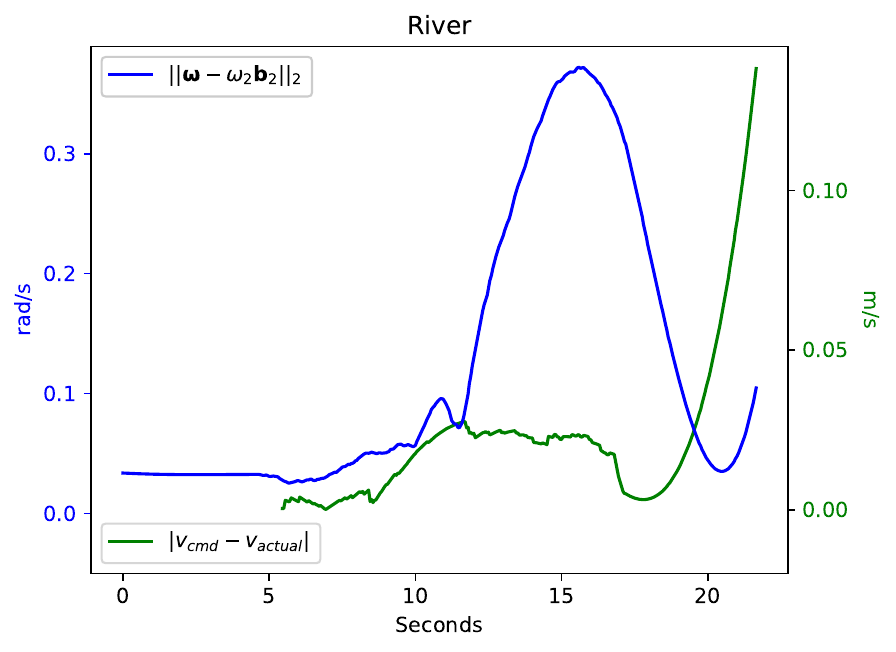}
    \includegraphics[width=.45\linewidth]{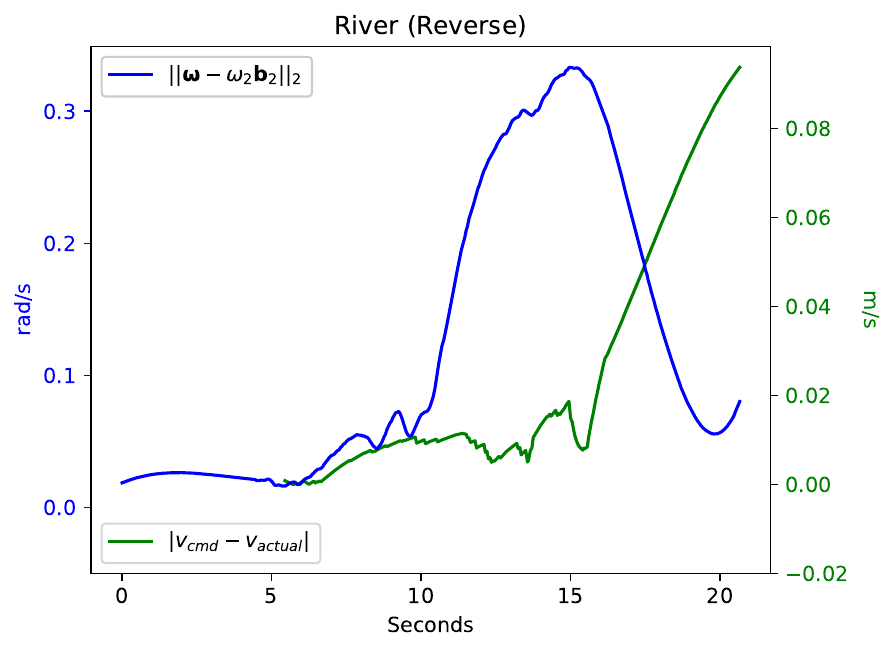}
    \includegraphics[width=.45\linewidth]{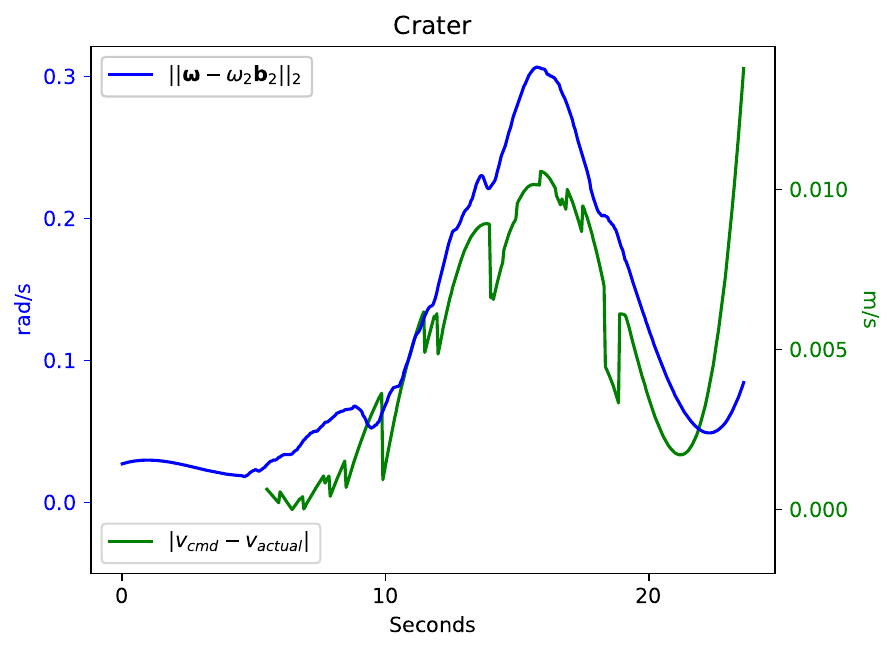}
    \includegraphics[width=.45\linewidth]{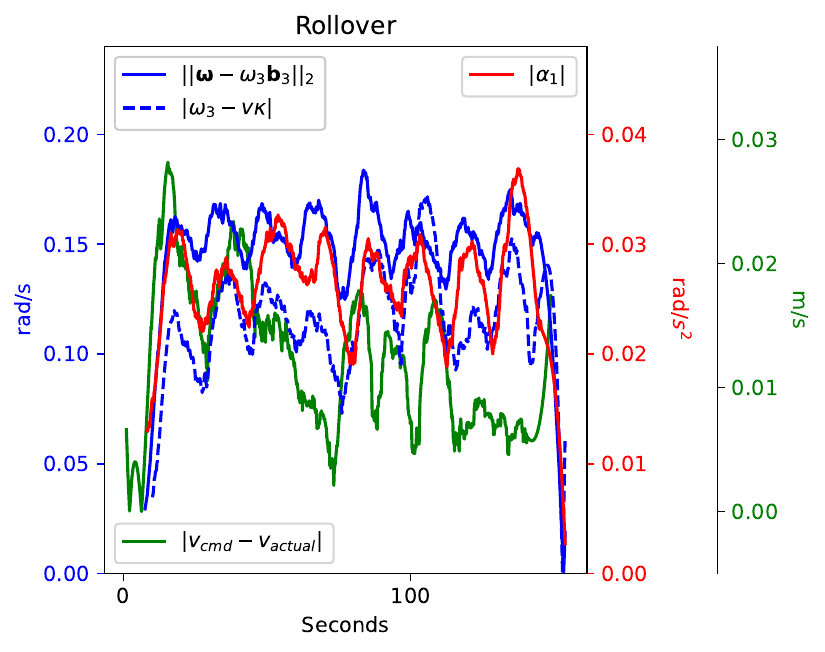}
    \caption{Plots of empirical constraint error terms for V-Ditch, River, Crater, and rollover experiments. See \Cref{tab:violations} for these errors' effects on the constraint values.}
    \label{fig:constraint_err_plots}
\end{figure*}

We plot the relevant states obtained using the IMU and visual-inertial odometry during our experiments in \cref{fig:constraint_err_plots} and summarize their largest empirical values in \Cref{tab:violations}.

\subsection{Project Website}
\urlstyle{tt}
Footage of experiments can be found on our project website here: 
\url{https://sites.google.com/cs.washington.edu/off-road-mpc}
.

\end{document}